\documentclass{article}
\usepackage{amsmath,amssymb,amsfonts}
\usepackage{cim2026}
\usepackage{times}
\usepackage{ifpdf}
\usepackage[english]{babel}
\usepackage{booktabs}
\usepackage{enumitem}
\usepackage{microtype}
\usepackage{graphicx}
\usepackage{subcaption}
\usepackage{tikz}
\usetikzlibrary{
  arrows.meta,
  positioning,
  calc,
  decorations.pathreplacing,
  shapes.geometric,
  fit,
  backgrounds,
  matrix
}

\def\papertitle{FROM LOCAL KERNELS TO GLOBAL FORM: MODELING THE EMERGENCE OF MUSICAL CONTENT}
\def\firstauthor{Francesco Vitucci}
\def\secondauthor{Michele Lorusso}
\def\thirdauthor{Francesco Scagliola}

\ifpdf
  \usepackage[pdftex,
    pdftitle={\papertitle},
    pdfauthor={Francesco Vitucci, Michele Lorusso, Francesco Scagliola},
    bookmarksnumbered,
    pdfstartview=XYZ
   ]{hyperref}
\else
  \usepackage[dvips,
    bookmarksnumbered,
    pdfstartview=XYZ
  ]{hyperref}
\fi

\hypersetup{
    colorlinks,
    citecolor=black,
    filecolor=black,
    linkcolor=black,
    urlcolor=black
}

\title{\papertitle}

\threeauthors
  {\firstauthor} {Conservatorio di Musica \\ ``N. Piccinni'' di Bari \\ {\scriptsize\tt francescovitucci1@gmail.com}}
  {\secondauthor}{Conservatorio di Musica \\ ``N. Piccinni'' di Bari \\ {\scriptsize\tt michelelorusso99@outlook.com}}
  {\thirdauthor} {Conservatorio di Musica \\ ``N. Piccinni'' di Bari \\ {\scriptsize\tt francesco.scagliola@gmail.com}}

\begin{document}

\maketitle

\abstract
Markov models are established tools for symbolic music, including non-homogeneous
formulations. The narrower contribution examined here is an observation-driven estimation
mechanism: overlapping sliding windows derive a trajectory of local transition kernels from
one symbolic sequence rather than from an exogenous formal partition. We test this mechanism
on 273 logical note events from Debussy's \textit{Syrinx} (1913), using the often-proposed
A--B--A' reading as a reference rather than ground truth. We apply the same validation to
absolute-pitch and notated-duration kernels. At $L=6$, both reference boundaries attain
the Jensen--Shannon maximum in both dimensions; the duration plateau is substantially
narrower (64 of 267 comparisons) than the pitch plateau (210 of 267). Because the theoretical maximum for consecutive sliding-window comparisons is set by window geometry and equals $1/\sqrt{L-1}$ for maximal turnover of the entering/leaving transition, the pitch value at $L=6$ and its broad plateau are not, by themselves, strong evidence. Their cross-dimensional alignment is consistent with boundary sensitivity, while the broad plateaus
preclude treating either curve alone as a unique automatic segmenter. Five-hundred-draw
re-synthesis experiments quantify departure from the source in both dimensions and expose
an exact-copy degeneracy at $L=2$.
\endabstract

\section{Introduction}\label{sec:introduction}

The use of Markov chains in algorithmic composition has a long tradition. From the
pioneering experiments of the 1950s and 1960s to contemporary applications in style
imitation, real-time improvisation, and hybrid symbolic/audio modelling
\cite{Ames1989, Nierhaus2009, Fernandez2013}, Markov processes have provided a
mathematically transparent framework for capturing local statistical regularities in
musical sequences. A transition matrix can be estimated from small corpora, inspected
directly, and reasoned about in musical terms.

First-order memorylessness and time-homogeneity are distinct assumptions. This paper
retains the former and relaxes the latter: transition probabilities may change with
position while each next state still depends only on the current state. Ames already
identified ``evolving transition matrices'' as a useful compositional extension
\cite{Ames1989}, and musical non-homogeneous Markov models are not new
\cite{Buenger2012,Roig2018}. Our specific object of study is therefore not
non-homogeneity itself, sliding-window estimation, or local statistics in general, but the
combination and evaluation of directed conditional-transition structures derived from one
symbolic sequence: local transition kernels obtained directly from the data, their temporal
trajectory, and analytical and generative uses of that same representation.

Windowed estimation is a standard response to non-stationarity in time-series and
data-stream analysis \cite{Gama2014}. In music, Chawin and Rom apply sliding-window
pitch-class histograms to form labelling \cite{Chawin2021}. We ask what changes when
the local statistic is instead a directed conditional-transition structure, and whether
that structure supplies useful analytical or generative evidence in a single-piece case
study.

Section~2 gives the formalism, Section~3 motivates local estimation,
Section~4 defines the estimator, Section~5 reports the tests on \textit{Syrinx}, and
Sections~6--8 discuss scope, related work, and conclusions.

\section{Background: Markov Models}\label{sec:background}

\subsection{Markov Chains}

Let $S = \{s_1, \dots, s_n\}$ be a finite state space. A first-order Markov chain
$\{X_t\}_{t \geq 1}$ satisfies \cite{Markov1906, Douc2018}:
\begin{equation}
\begin{aligned}
&\mathbb{P}(X_{t+1} = s_j \mid X_t = s_i, X_{t-1}, \dots)\\
&\qquad= \mathbb{P}(X_{t+1} = s_j \mid X_t = s_i).
\end{aligned}
\label{eq:markov_property}
\end{equation}
A time-homogeneous chain is fully determined by $P \in \mathbb{R}^{n\times n}$,
where $P_{ij}:=\mathbb{P}(X_{t+1}=s_j\mid X_t=s_i)$ and $\sum_jP_{ij}=1$. In musical applications $S$ is typically a vocabulary of symbolic
events: pitches, pitch classes, chord labels, rhythmic durations, or composite
tokens \cite{Ames1989, Nierhaus2009}.

\subsection{Empirical Estimation}

In practice $P$ must be estimated from an observed sequence $(x_1, \dots, x_T)$.
The standard estimator counts transitions $C_{ij} = \#\{t \mid x_t = i,\, x_{t+1} = j\}$
and normalises each row:
\begin{equation}
P_{ij} =
\begin{cases}
C_{ij} \,/\, {\textstyle\sum_k C_{ik}} & \text{if } \sum_k C_{ik} > 0, \\
0 & \text{otherwise.}
\end{cases}
\label{eq:empirical_estimator}
\end{equation}
In our implementation the sequence is augmented with $\langle\mathrm{START}\rangle$ and
$\langle\mathrm{END}\rangle$ boundary tokens, so that first and last events are explicitly
represented as transitions. The terminal row is a sub-Markovian kernel \cite{Douc2018}
with no outgoing mass; it is excluded from probability sampling in generative use.

\section{Why Estimate Local Kernels?}\label{sec:limits}

\subsection{Memorylessness and Musical Perception}

The Markov property and time-homogeneity are logically independent. A chain may remain
memoryless while its transition kernel changes with time; conversely, a time-homogeneous
finite-state chain need not have a unique invariant distribution, nor need its distributions
converge to one \cite{Douc2018, Meyn2009}. The limitation
addressed here is the combination of first-order memorylessness with a single globally
constant kernel. Under that combination, all long-range structural information is
compressed into one matrix.

The problem is not merely technical. Musical events are not perceived as isolated
symbols; they acquire meaning through their position in a temporal structure, through
recurrence, expectation, tension, and resolution. A model that treats every occurrence
of a state as statistically equivalent, regardless of its formal placement, discards
precisely the temporal context that makes a sequence intelligible as form. This is not
an argument against the Markov property as such, but against the further assumption
that the transition law itself is invariant over the entire duration of the piece.

In musical contexts this conflation is consequential: a single homogeneous matrix cannot
distinguish between transitions that occur in structurally distinct regions, even when
those regions are perceptually or syntactically dissimilar \cite{Ames1989, Buenger2012}.

\subsection{Higher-Order Models Are Not the Answer}

Higher-order chains model $\mathbb{P}(X_{t+1} \mid X_t, \dots, X_{t-m+1})$, enriching
local dependency. The IDyOM framework of Pearce and Wiggins employs variable-order
models with interpolated smoothing to predict listeners' melodic expectations
\cite{Pearce2005, PearceWiggins2012}, and is effective for that purpose.

However, if the model remains time-homogeneous, an identical subsequence appearing in
different formal regions is still treated as statistically equivalent, regardless of
its temporal position. Variable order captures \emph{depth} of context; our approach
captures \emph{location} in time. Higher-order models also entail rapid state-space
growth, leading to data sparsity in finite or windowed corpora. The issue does not lie
in the Markov property itself, but in requiring time-homogeneity from a sequence
whose syntactic law changes over time. As established in the theory of
time-inhomogeneous chains \cite{Douc2018, Levin2017}, a process may remain Markovian
while its transition kernel varies with time.

\section{Time-Varying Markov Model via Sliding Windows}\label{sec:model}

Following standard windowed estimation \cite{Douc2018,Gama2014}, we construct
a sequence of local matrices,
\[
P^{(1)},\; P^{(2)},\; \dots,\; P^{(W)},
\]
each associated with a short temporal segment of the symbolic sequence.

\subsection{Sliding Window Construction}

\begin{figure}[h]
\centering
\resizebox{\columnwidth}{!}{%
\begin{tikzpicture}[
  >=Stealth,
  symbol/.style={draw,thick,rectangle,minimum size=0.72cm,
                 font=\scriptsize,inner sep=1pt},
  wbox/.style={rounded corners=2pt,thick,fill opacity=0.35,draw opacity=1},
  matnode/.style={draw,thick,minimum width=0.85cm,minimum height=0.60cm,
                  inner sep=2pt,font=\scriptsize},
  font=\scriptsize
]
  \foreach \i in {1,...,9} {
    \node[symbol] (s\i) at (\i*0.82, 0) {$x_{\i}$};
  }
  \node at (10*0.82 + 0.1, 0) {$\cdots$};
  \foreach \i in {1,...,6} {
    \draw[densely dotted,gray!50] (\i*0.82, -0.39) -- (\i*0.82, -2.80);
  }
  \node[blue!60!black,font=\scriptsize\bfseries] at (-0.12, -1.00) {$w_1$};
  \draw[wbox, draw=blue!70!black, fill=blue!30]
    (1*0.82-0.39, -1.20) rectangle (4*0.82+0.39, -0.80);
  \draw[->,thick,blue!70!black] (4*0.82 + 0.50, -1.00) -- (8.70, -1.00);
  \node[matnode, fill=blue!18] at (9.35, -1.00) {$P^{(1)}$};
  \node[red!60!black,font=\scriptsize\bfseries] at (-0.12, -1.78) {$w_2$};
  \draw[wbox, draw=red!70!black, fill=red!30]
    (2*0.82-0.39, -1.98) rectangle (5*0.82+0.39, -1.58);
  \draw[->,thick,red!70!black] (5*0.82 + 0.50, -1.78) -- (8.70, -1.78);
  \node[matnode, fill=red!18] at (9.35, -1.78) {$P^{(2)}$};
  \node[green!45!black,font=\scriptsize\bfseries] at (-0.12, -2.56) {$w_3$};
  \draw[wbox, draw=green!50!black, fill=green!35]
    (3*0.82-0.39, -2.76) rectangle (6*0.82+0.39, -2.36);
  \draw[->,thick,green!50!black] (6*0.82 + 0.50, -2.56) -- (8.70, -2.56);
  \node[matnode, fill=green!18] at (9.35, -2.56) {$P^{(3)}$};
  \node[gray!70,font=\scriptsize] at (-0.12, -3.20) {$\vdots$};
  \node[gray!70,font=\scriptsize] at (9.35,  -3.20) {$\vdots$};
\end{tikzpicture}%
}
\caption{Sliding window ($L=4$, step $=1$). Each row is one window position; from
every $w_m$ a local transition matrix $P^{(m)}$ is estimated. Successive windows overlap
by $L{-}1$ symbols, so the estimated kernel changes smoothly as the window advances.}
\label{fig:sliding-window}
\end{figure}
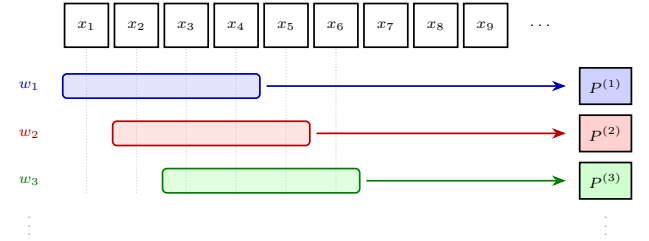

Given a window length $L$ and step size $1$, define overlapping windows
$w_m = (x_m, \dots, x_{m+L-1})$ for $m = 1, \dots, W = T-L+1$.
Each window contains $L-1$ transitions, indexed by
$I_m = \{m, \dots, m+L-2\}$, yielding a local count matrix
$C^{(m)}_{ij} = \#\{r \in I_m \mid x_r = i,\, x_{r+1} = j\}$.
Figure~\ref{fig:sliding-window} illustrates the construction geometrically.

Each $\widehat{P}^{(m)}$ is obtained by applying Eq.~\eqref{eq:empirical_estimator}
to $C^{(m)}$, acting as a locally stationary kernel under the working assumption that
transition probabilities are approximately constant within the window. The choice of $L$
determines a bias--variance trade-off: large windows reduce estimation variance but blur
local change; small windows track change at the cost of sparse rows. For optional
shrinkage toward the global estimate $P^{(G)}$, we use
\begin{equation}
\widetilde P^{(m)}_{ij}=
\frac{C^{(m)}_{ij}+\lambda P^{(G)}_{ij}}
{\sum_k C^{(m)}_{ik}+\lambda},\qquad \lambda>0.
\label{eq:shrinkage}
\end{equation}
The reported baseline instead uses unsmoothed counts and backs off to $P^{(G)}$ only
when the selected local row is empty. Thus the observed matrix $\widehat P^{(m)}$ is
not, by itself, a complete stochastic kernel. The kernel actually used for generation is
defined row-wise by
\begin{equation}
K^{(m)}_{ij}=
\begin{cases}
\widehat P^{(m)}_{ij}, & \text{if local row }i\text{ has positive mass},\\
P^{(G)}_{ij}, & \text{otherwise}.
\end{cases}
\label{eq:generation_kernel}
\end{equation}
Boundary tokens are masked in analytical comparisons.

\subsection{Time-Varying Transition Kernels}

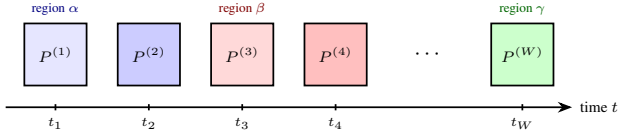
\begin{figure}[h]
\centering
\resizebox{\columnwidth}{!}{%
\begin{tikzpicture}[
  >=Stealth,
  mat/.style={draw,thick,minimum width=1.0cm,minimum height=1.0cm,
              font=\scriptsize,inner sep=2pt}
]
  \node[mat,fill=blue!10]  (P1) at (0,0)   {$P^{(1)}$};
  \node[mat,fill=blue!20]  (P2) at (1.5,0) {$P^{(2)}$};
  \node[mat,fill=red!15]   (P3) at (3.0,0) {$P^{(3)}$};
  \node[mat,fill=red!25]   (P4) at (4.5,0) {$P^{(4)}$};
  \node[font=\normalsize]      at (6.0,0)  {$\cdots$};
  \node[mat,fill=green!20] (PW) at (7.5,0) {$P^{(W)}$};
  \draw[->,thick] (-0.8,-0.85) -- (8.3,-0.85)
    node[right,font=\scriptsize] {time $t$};
  \foreach \x/\name in {0/$t_1$,1.5/$t_2$,3.0/$t_3$,4.5/$t_4$,7.5/$t_W$} {
    \draw[thick] (\x,-0.80) -- (\x,-0.90);
    \node[below,font=\tiny] at (\x,-0.90) {\name};
  }
  \node[above,font=\tiny,blue!50!black]  at (P1.north) {region $\alpha$};
  \node[above,font=\tiny,red!50!black]   at (P3.north) {region $\beta$};
  \node[above,font=\tiny,green!40!black] at (PW.north) {region $\gamma$};
\end{tikzpicture}%
}
\caption{Trajectory of transition matrices. Colour-coding by formal region shows how
the estimated kernel evolves as the sequence moves from one section to another.}
\label{fig:trajectory}
\end{figure}

The output is a \emph{trajectory of matrices} (Figure~\ref{fig:trajectory}):
$\mathcal{P} = \{\widehat{P}^{(m)}\}_{m=1}^{W}$.
This collection admits three complementary readings: a local description of stylistic
behaviour; a temporal evolution of musical grammar; and a piecewise-constant approximation
of an underlying time-inhomogeneous Markov process.

\begin{figure}[h]
  \centering
  \includegraphics[width=\columnwidth]{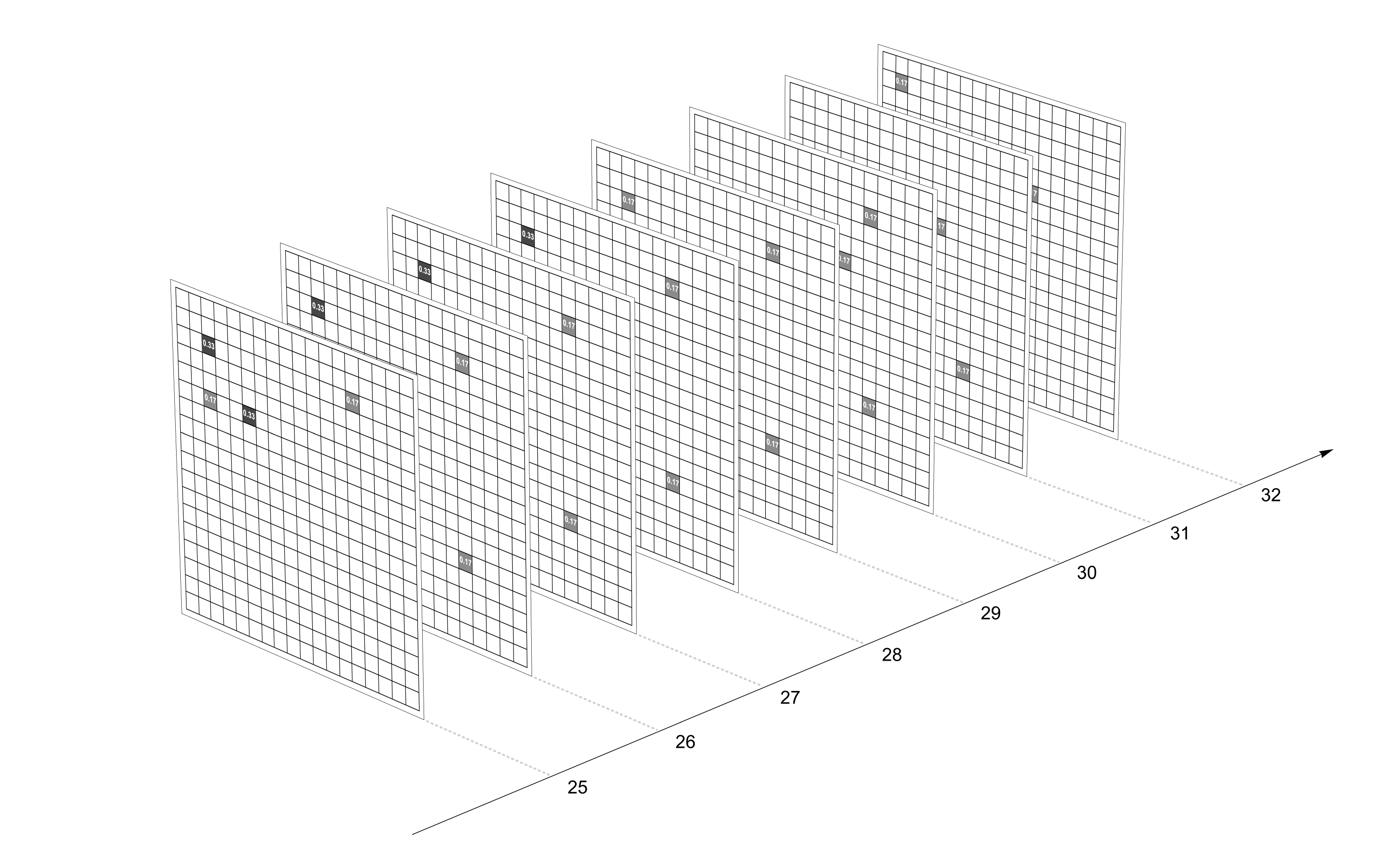}
  \caption{Eight successive matrices from the trajectory ($L=6$, $t=25,\ldots,32$).
  Darker cells indicate higher transition probability. With only five transitions per
  window, the support is sparse and changes whenever one transition enters or leaves.}
  \label{fig:matrixStack}
\end{figure}

The approach does not violate the Markov property: conditional independence from the
past is preserved at each step. What is relaxed is time-homogeneity. An empirical
example of the resulting trajectory is shown in Figure~\ref{fig:matrixStack}.

\subsection{Mathematical Formalisation}

To use the trajectory generatively, a schedule $g(r)$ maps each transition time $r$
to one local estimate, yielding a time-inhomogeneous Markov chain \cite{Douc2018}:
\begin{equation}
\mathbb{P}(X_{r+1} = j \mid X_r = i) = \widehat{P}^{(g(r))}_{ij}.
\end{equation}
Since windows overlap, $g$ is not unique; first, last, or centre-aligned choices all
define valid chains but are not equivalent models. In the re-synthesis below we use
centre alignment, backing off to the global kernel when a local row has zero mass.

For generative use we instantiate the estimator on two aligned symbolic streams.
Absolute MIDI pitch yields $\mathcal{P}$; exact notated duration yields
$\mathcal{D}=\{\widehat{D}^{(m)}\}_{m=1}^{W}$. The shared schedule selects the complete
generation kernels $K^{(g(r))}$ and $K_D^{(g(r))}$, obtained by the same row-wise fallback
construction, while the two kernels remain separately
inspectable and sampled rather than forming one joint pitch--rhythm state space. In particular,
the pitch chain is defined by
\begin{equation}
\Pr(X_{r+1}=j\mid X_r=i)=K^{(g(r))}_{ij},
\label{eq:inhomogeneous_generation}
\end{equation}
and analogously for duration using $K_D^{(g(r))}$, not the incomplete observed matrix
$\widehat P^{(g(r))}$.

\section{Case Study: Debussy's \textit{Syrinx}}\label{sec:casestudy}

\subsection{Choice of Piece}

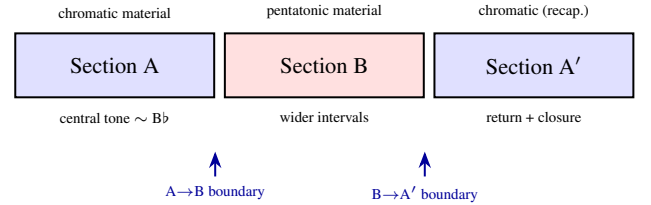
\begin{figure}[h]
\centering
\resizebox{\columnwidth}{!}{%
\begin{tikzpicture}[
  >=Stealth,
  sect/.style={draw,thick,minimum height=0.9cm,minimum width=2.8cm,font=\small},
]
  \node[sect,fill=blue!12] (A)  at (0,0)    {Section A};
  \node[sect,fill=red!12]  (B)  at (2.95,0) {Section B};
  \node[sect,fill=blue!12] (Ap) at (5.90,0) {Section A$'$};
  \node[above=2pt of A, font=\tiny]  {chromatic material};
  \node[above=2pt of B, font=\tiny]  {pentatonic material};
  \node[above=2pt of Ap,font=\tiny]  {chromatic (recap.)};
  \node[below=2pt of A, font=\tiny]  {central tone $\sim$ B$\flat$};
  \node[below=2pt of B, font=\tiny]  {wider intervals};
  \node[below=2pt of Ap,font=\tiny]  {return + closure};
  \draw[<-,thick,blue!60!black]
    ($(A.south east)+(0,-0.75)$) -- ($(A.south east)+(0,-1.10)$)
    node[below,font=\tiny] {A$\to$B boundary};
  \draw[<-,thick,blue!60!black]
    ($(B.south east)+(0,-0.75)$) -- ($(B.south east)+(0,-1.10)$)
    node[below,font=\tiny] {B$\to$A$'$ boundary};
\end{tikzpicture}%
}
\caption{Working A--B--A' reference for \textit{Syrinx}. The two sectional
coordinates are comparison targets, not ground-truth labels.}
\label{fig:syrinx-form}
\end{figure}

Debussy's \textit{Syrinx} (1913), for solo flute, is brief, strictly monodic, and
built on a small number of recognisable melodic ideas, making the symbolic sequence
manageable and analytically transparent. Its macroformal organisation is often read as tripartite A--B--A'
\cite{Nattiez1975,Nattiez1990}. Curinga's synoptic comparison shows broad convergence on
that macroform but substantial disagreement over internal divisions; measures 9 and 26
recur across several published readings \cite{Curinga2001}. Leplat explicitly labels
mm.~1--8 as A, 9--25 as B, and 26--35 as A' \cite[pp.~9--10]{Leplat2024}.
We therefore use the beginnings of measures 9 and 26 (positions 6 and 18.75 in whole-note
units) as falsifiable references, not ground-truth labels. Figure~\ref{fig:syrinx-form}
summarises this working plan.

\subsection{Symbolic Representation and Parameters}

The one-part MusicXML score is imported as symbolic note, rest, and state entities.
For analysis, logical note attacks are ordered by score position; absolute pitch and exact
notated duration are retained without conversion to performance seconds. START and END
tokens are not used in the distance calculations.

We compute kernels for $L\in\{2,6,20\}$, containing one, five, and nineteen
transitions. These three values were fixed before the quantitative comparisons. The qualitative
assessment of the window range had identified $L=6$ as a particularly interpretable balance;
the quantitative comparison is therefore not an independent selection of $L=6$. The MusicXML
score contains 308 written note entries. We exclude 14
zero-duration grace notes and merge 21 tied continuations into their preceding attacks,
yielding two aligned 273-event streams: absolute MIDI pitch (28 states) and exact notated
duration after tie merging (23 rational states). Both streams use the same windows and
fixed score-based reference coordinates, but are evaluated separately.

\subsection{Successive transition-distribution distance test}

\begin{figure*}[t]
  \centering
  \includegraphics[width=0.96\textwidth]{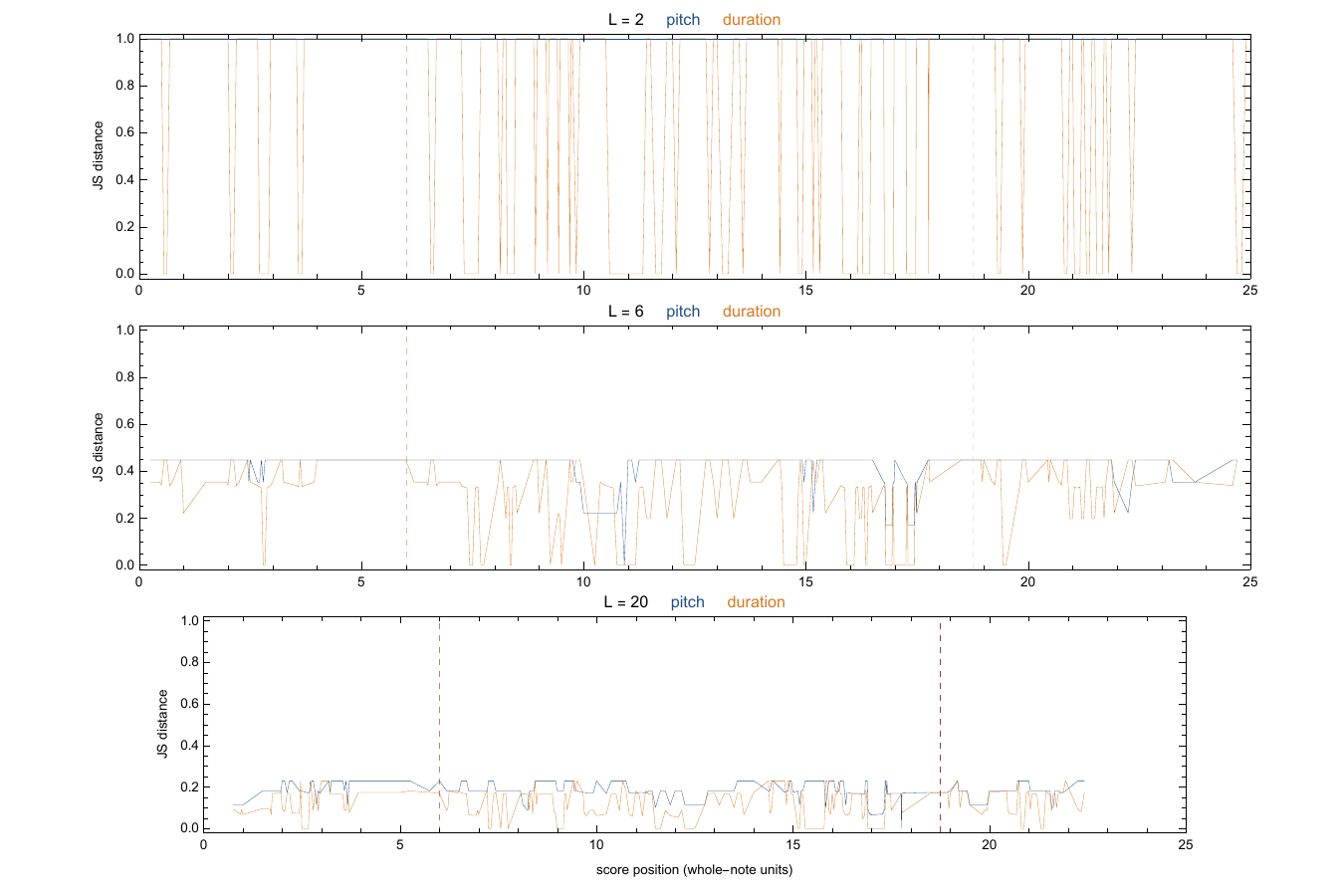}
  \caption{Successive transition-distribution JS distance between absolute-pitch and
  notated-duration transition distributions. Dashed lines mark the measure-9 and measure-26
  references. A point
  on a high plateau is not an isolated boundary peak.}
  \label{fig:successive-js}
\end{figure*}

For each event stream and $L$, we vectorise and normalise each local transition-count matrix
as an empirical joint distribution over transition pairs $(i,j)$. We compare consecutive
windows using Jensen--Shannon (JS) and Hellinger successive transition-distribution distances;
Frobenius is reported as a row-normalised kernel distance. Forward KL
is infinite in 223 pitch and 125 duration comparisons out of 267 at $L=6$ because mass
disappears from the support, so it is not used as a peak curve.
Figure~\ref{fig:successive-js} shows the unsmoothed JS curves.

The curves provide scale-dependent observations whose interpretation is constrained by the
geometry of consecutive sliding windows. For maximal turnover of the single transition that
enters and leaves, the theoretical observed maximum is $1/\sqrt{L-1}$; hence the maximum 0.447
at $L=6$ is the geometrically expected ceiling, not independent evidence of an unusually strong
boundary effect. At $L=2$, pitch has 271 and
duration 152 maximum-valued comparisons, so the scale is non-selective. At $L=6$, both
references attain the maximum 0.447 in both streams. The plateau contains 210 pitch
comparisons (78.7\%) but only 64 duration comparisons (24.0\%): rhythmic transition
structure supplies markedly greater selectivity, while its agreement with pitch at both
references is consistent with the cited A--B--A' articulation. At $L=20$, evidence becomes
asymmetric: pitch yields 0.229 and 0.174 at the references, with the first value maximal
but shared by 74 comparisons; duration yields 0.170 and 0.174, both below its 0.229 maximum,
shared by 20 comparisons. The pre-specified three-value comparison therefore locates the strongest
cross-dimensional boundary alignment at $L=6$, which is consistent with boundary sensitivity
but does not turn the curves into a unique
automatic segmentation.

\subsection{Generative Re-synthesis}

We evaluate centre-aligned, hard-backoff pitch and duration samplers over 500
deterministic seeds. Every draw starts from its source state and has 273 events. For each
stream we report aligned state mismatch, whole-sequence transition JS, and mean JS computed
separately inside the three reference regions (Table~\ref{tab:resynthesis}).

\begin{table}[h]
\centering
\caption{Re-synthesis departure: mean [2.5\%, 97.5\%], 500 draws; P=absolute pitch, D=notated duration.}
\label{tab:resynthesis}
\scriptsize
\setlength{\tabcolsep}{3pt}
\begin{tabular}{@{}ccccc@{}}
\toprule
stream & $L$ & mismatch & global JS & sectional JS \\
\midrule
P & 2  & 0 [0, 0] & 0 [0, 0] & 0 [0, 0] \\
P & 6  & .678 [.462, .912] & .167 [.116, .213] & .196 [.128, .255] \\
P & 20 & .839 [.711, .945] & .303 [.251, .358] & .366 [.299, .431] \\
D & 2  & 0 [0, 0] & 0 [0, 0] & 0 [0, 0] \\
D & 6  & .344 [.293, .392] & .129 [.091, .172] & .177 [.126, .231] \\
D & 20 & .588 [.509, .663] & .222 [.175, .276] & .320 [.249, .397] \\
\bottomrule
\end{tabular}
\end{table}

With $L=2$, each aligned local row contains only the observed next transition, so both
samplers copy their source exactly. At $L=6$ and $L=20$, both dimensions depart substantially
at event level despite lower transition distances. The parallel evaluation therefore tests
rhythmic departure as well as melodic departure, but not their joint coherence: separate
sampling assumes conditional independence of pitch and duration given formal position.
The plotted outputs remain illustrative draws, not evaluation evidence.

\begin{figure}[h]
  \centering
  \includegraphics[width=\columnwidth]{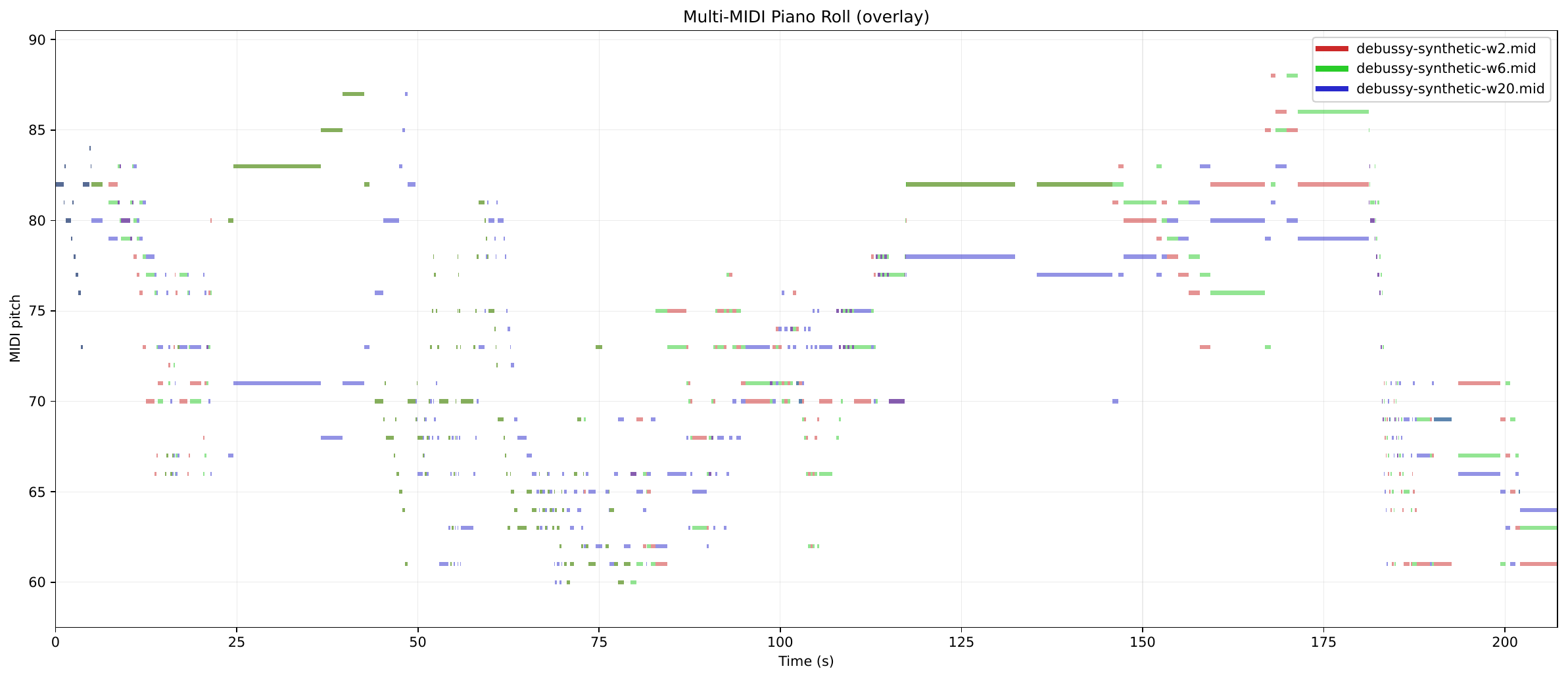}
  \caption{One illustrative draw for each $L$. Quantitative results are reported in
  Table~\ref{tab:resynthesis}; visual similarity is not used to select a window.}
  \label{fig:resynth3}
\end{figure}

A shrinkage pilot using Eq.~\eqref{eq:shrinkage} with $\lambda=1$ removes the
$L=2$ exact-copy degeneracy, but is not neutral. At $L=6$, pitch mismatch rises from
0.678 to 0.881 and sectional JS from 0.196 to 0.618; duration mismatch rises from 0.344
to 0.672 and sectional JS from 0.177 to 0.476. We retain hard backoff as the explicit
baseline and leave shrinkage strength to held-out selection on a larger corpus.

The event-level tables, complete kernel-distance trajectories, re-synthesis summaries,
and extended figures supporting this case study are archived in the companion Zenodo
record \cite{Vitucci2026ExtendedMaterials}.

\section{Discussion}\label{sec:applications}

The distance test supports a bounded case-study claim while defining its scope. At $L=6$,
both score-based references coincide with maximal change in pitch and duration. Since the
maximum is constrained by the sliding-window geometry, this coincidence is consistent with
boundary sensitivity rather than a particularly strong validation of it; the
rhythmic curve is appreciably more selective. This is boundary-aligned evidence from two
complementary event dimensions. Broad plateaus nevertheless show that descriptive
validation is not equivalent to a turnkey segmentation algorithm. That distinction is
musically appropriate: published analyses converge on a tripartite macroform while proposing
different internal partitions \cite{Curinga2001}. The sensitivity of candidate counts to
comparison scale likewise favours a multiscale reading over a single mandatory partition.
A corpus study must predefine comparison lag, regularisation, a cross-dimensional rule,
peak criterion, and reference tolerance.

The present distance compares strongly overlapping consecutive windows, in which only one
transition exits and one enters. It is therefore sensitive to local turnover, but this geometry
limits its interpretation as a true boundary detector. Future work should compare
boundary-centred preceding and following windows, preferably non-overlapping, through a measure
such as $D(P_t^-,P_t^+)$.

Generatively, explicit local kernels remain attractive because a composer can inspect,
edit, or reschedule them. Yet formal control cannot be inferred from a single piano roll or
from global distributional similarity. Evaluation should distinguish memorisation,
event-level departure, within-section syntax, cross-dimensional pitch--duration coherence,
and perceptual judgement.

\section{Related Work}\label{sec:related}

Markov chains have been widely used in algorithmic composition and musical
analysis \cite{Ames1989, Nierhaus2009, Shapiro2021, Temperley2007}. Most approaches
rely on homogeneous models. For example, Catak et al.\ estimate note-transition
probabilities from 25 classical scores and compare Markov-chain generation with an
RNN \cite{Catak2021}. More recently, Pleshkova and Kostov derive separate transition
matrices from genre-specific chord progressions for guitar-melody generation
\cite{Pleshkova2025}; these matrices are conditioned by corpus category and chord context
rather than by local position within a single piece. Non-homogeneous formulations include Buenger's phrase-based
model \cite{Buenger2012}, where non-homogeneity is tied to a hand-specified phrase
partition, and Roig et al.'s beat-based harmony model \cite{Roig2018}, where transition
probabilities depend on position within a recurring metrical grid. Both differ from our
approach in that non-homogeneity is defined exogenously rather than extracted from the
data by windowing. Esqu{\'i}vel et al.\ propose an estimation--calibration procedure for
continuous-time non-homogeneous Markov chains with finite state spaces in a non-musical
setting \cite{Esquivel2024}. In discrete-time financial forecasting, Wili{\'n}ski estimates
changing first- and second-order transition matrices from sequences of fixed-length time
windows, optimizing the window and state-discretisation parameters \cite{Wilinski2019}.
Although observation-driven, the objective is next-state prediction rather than
time-resolved structural analysis of a single symbolic artefact.

Sliding-window estimation is standard in non-stationary data analysis
\cite{Gama2014} and common in MIR. Chawin and Rom show that local pitch-class histograms
improve sonata-form labels \cite{Chawin2021}; their distributions are marginal, whereas
ours encode directed conditional transitions. Thus the contribution is the application and
evaluation, in a single symbolic sequence, of window-derived directed conditional-transition
structures and their temporal trajectory for both analysis and generation, not sliding windows,
local statistics, or non-homogeneous chains per se. The IDyOM framework \cite{Pearce2005,
PearceWiggins2012} addresses depth of context via variable-order models; our approach
addresses temporal location. Both relaxations of the basic first-order homogeneous chain
are in principle combinable.

Recent surveys map the broader field of deep-learning-based symbolic music generation,
including representations, algorithms, evaluation methods, and open challenges
\cite{JiSurvey2023}. Transformer-based architectures achieve state-of-the-art results in symbolic music
generation \cite{Muhamed2021, Shih2022, Ji2024, Wang2024, Qu2024}, but require large
datasets and do not provide the interpretable, time-resolved transition matrices that
are central to the present approach. The two paradigms are complementary rather than
competing.

\section{Conclusion}\label{sec:conclusion}

Overlapping windows provide an interpretable, observation-driven way to estimate
changing transition kernels. In \textit{Syrinx}, both reference boundaries attain maximal
successive transition-distribution distance in both pitch and duration at $L=6$, with greater selectivity in the
rhythmic trajectory. Together with the published convergence on a tripartite macroform,
this is consistent with the central claim that local kernels can expose formal change
obscured by a single global estimator. Re-synthesis further demonstrates controllable
departure in both dimensions as $L$ changes. The result is a first-case, multiscale observation
consistent with boundary sensitivity,
not a claim of unique automatic segmentation or corpus-level generality.

\bibliography{references}

\end{document}